\documentclass[runningheads]{llncs}
\usepackage[T1]{fontenc}
\usepackage{placeins}

\usepackage{cite}
\usepackage{comment}
\usepackage{caption}
\usepackage{amsmath}
\usepackage{algorithm}
\usepackage{amssymb}
\usepackage{algpseudocode}
\usepackage{orcidlink}
\usepackage{graphicx}
\usepackage{booktabs}
\usepackage{siunitx}
\usepackage{subfigure}

\begin{document}
\title{An Empirical Study of Feature Selection Granularity}


\author{Muhammad Rajabinasab \inst{1}\orcidID{0009-0006-7045-3998} \and
Arthur Zimek \inst{1}\orcidID{0000-0001-7713-4208}}

\authorrunning{M. Rajabinasab \& A. Zimek}

\institute{Department of Mahematics and Computer Science\\ University of Southern Denmark, Odense, Denmark\\
\email{\{rajabinasab, zimek\}@imada.sdu.dk}
}

\maketitle              
\begin{abstract}
Feature selection aims to identify the most informative and relevant features for a given dataset, either in terms of capturing the underlying data structure and distribution better, or with respect to the performance on a downstream task. Existing research in this area has largely focused on developing novel algorithms (in both supervised and unsupervised settings), proposing new evaluation metrics and frameworks, or benchmarking the performance of existing methods. In this work, we examine feature selection through an algorithmic design perspective. Conventional feature selection algorithms typically compute feature importance scores globally across the entire feature set and then select the top-ranked features in a single step. However, this approach raises a critical question: Can the presence of less informative (or noisy) features mask or obscure the true importance of other, more relevant features? In other words, would a recursive strategy, where features are removed one by one while re-evaluating importance at each step, yield different and potentially better results than the standard global ranking approach? To answer this question, we conduct an extensive empirical study using five diverse feature selection algorithms. We implement each algorithm under both the conventional global selection design and the greedy recursive elimination design. We then analyze the impact of this algorithmic choice, both individually for each method and collectively across all methods, on a range of standard feature selection evaluation metrics. The empirical evaluation results show that the greedy approach improves the overall feature selection quality almost consistently, albeit on the expense of higher computational cost, supporting our initial expectation that the curse of dimensionality also obscures the ways of mitigating it. 

\end{abstract}
\section{Introduction}
The curse of dimensionality represents one of the most fundamental challenges in machine learning and data mining. It posits that, as dimensionality increases, we encounter both increased computational complexity inflicted by the excessive number of dimensions and a degradation in the performance of downstream tasks. For instance, the distances become less discriminative as dimensionality grows \cite{DBLP:journals/tkde/FrancoisWV07}, leading to the obscuration of various tasks such as Nearest Neighbors search \cite{RAJABINASAB2025126254}, classification \cite{DBLP:journals/tifs/AmsalegBBEFHRN21}, clustering \cite{DBLP:conf/sisap/OkkelsTAZS25}, outlier detection \cite{DBLP:conf/sdm/Anderberg0CHMRZ24, rajabinasab2026randomizedpcaforestunsupervised, rajabinasab2025hdsvdd}, and beyond. To address the curse of dimensionality, dimensionality reduction techniques are employed. These techniques encompass several distinct families based on the target application. For instance, in image processing applications such as face recognition, feature extraction is often used, as using the raw pixel data does not yield satisfactory recognition accuracy~\cite{asadi2021face}. One prominent family of dimensionality reduction is Feature Selection, which is particularly significant as it preserves the original meaning of each dimension and, consequently, maintains explainability \cite{guyon2003introduction}.

Feature Selection is the identification of the most informative features, whether with respect to the ability of capturing the underlying data structure and distribution or with regards to the performance of a downstream task \cite{guyon2003introduction}. Feature selection is done in both supervised and unsupervised fashion. In the supervised settings, feature selection is based on some properties of features directly (information w.r.t. class labels) or indirectly (assessing the performance of a classifier on selected features), while in the unsupervised setting, individual feature properties (e.g., variance) or pairwise feature relations (e.g.,  correlation) are used to calculate feature importance values. While much research has been done in the field of feature selection, the focus was mostly on proposing feature selection methods \cite{kononenko1994estimating, breiman2001random, chen2016xgboost, tibshirani1996regression}, evaluation frameworks and metrics \cite{rajabinasab2025metrics, rajabinasab2025fsdem}, or comparative surveys and benchmark studies \cite{li2017feature}.

Feature selection methods typically operate through the following sequential steps: i) A dataset consisting of $n$ instances and $d$ dimensions is provided to the feature selection algorithm. ii) A feature importance value is calculated based on a specific criterion that reflects the underlying philosophy of the method (e.g., the magnitude of Mutual Information between a feature and the labels). iii) These importance values are subsequently utilized to rank the features. iv) The top $d'$ features are selected, while the remaining features, deemed redundant or irrelevant, are discarded. This process yields a lower-dimensional, explainable subspace where $d' \ll d$, thereby addressing the curse of dimensionality. 
In some cases~\cite{guyon2002gene, asadi2021face}, redundant features are eliminated one by one, and the criterion is reevaluated until the desired dimensionality is reached.
A feature selection algorithm is considered effective if it significantly reduces dimensionality while preserving the performance of the downstream task. However, in many scenarios, particularly those involving high-dimensional data where $d$ is very large, removing features can actually lead to an improvement in the quality of downstream tasks. This occurs because redundant or irrelevant features often introduce masking effects, such as noise, which can obscure the underlying data structure and hinder model performance \cite{guyon2003introduction}. 

An important question arises: if redundant features are potent enough to obscure an entire dataset and degrade downstream task performance despite the presence of informative features, can they not also obscure the underlying logic of feature selection algorithms? Consider the execution of a feature selection algorithm using two distinct strategies: a standard global feature importance calculation versus an iterative greedy strategy where the least important, or most redundant, features are iteratively removed and the feature importance values are re-evaluated. A critical inquiry is whether these two approaches, applying the same feature selection criterion, yield divergent results and, if so, which strategy demonstrates superior performance.

In this paper, we provide an extensive empirical study to address this critical question. We select and evaluate five distinct feature selection algorithms, implemented using both, the global and the greedy~\cite{DBLP:journals/ai/KohaviJ97, DBLP:conf/icmla/EscanillaHKKSP18} strategy. To ensure a robust analysis, we conduct extensive experiments across 38 datasets, utilizing a comprehensive suite of evaluation metrics to elucidate the similarities and differences between both approaches. The evaluation metrics employed in this study reflect feature subset similarity, predictive performance (across both supervised and unsupervised tasks), model-agnostic quality assessment, and the scalability of the respective approaches.

The remainder of this paper is structured as follows: Given the absence of prior studies investigating the granularity of the feature selection process, Section~\ref{sec:fs} formally discusses the feature selection process and provides a review of the specific algorithms utilized in our study. Section~\ref{sec:meth} details the formal problem definition and the methodology underlying our empirical study to compare the standard global feature selection process with the iterative greedy approach. Section~\ref{sec:expset} outlines the experimental setup, including descriptions of the datasets, evaluation metrics, and the overall evaluation procedure. Section~\ref{sec:resu} presents the experimental results, accompanied by detailed analyses, investigations, and discussions designed to address the primary research question of this work. Finally, Section~\ref{sec:conc} provides concluding remarks and final discussions.

\section{Feature Selection} \label{sec:fs}
Feature selection is a dimensionality reduction technique which aims to reduce the dimensionality by removing redundant and irrelevant features. Let $\mathcal{F} = \{f_1, f_2, \dots, f_d\}$ denote the complete set of $d$ input features. The objective of feature selection is to identify a subset $S \subset \mathcal{F}$ that maximizes a scoring criterion $J(S)$, such as predictive power or information gain, while minimizing the cardinality of $S$. Formally, we seek:
\begin{equation}
    S^* = \arg \max_{S \subset \mathcal{F}} \{ J(S) - \lambda |S| \}
\end{equation}
where $\lambda$ is a penalty term for model complexity.

\subsection{Selected Feature Selection Methods}
Several methodologies exist within the domain of feature selection to mitigate the curse of dimensionality. For this empirical investigation, we employ five distinct feature selection techniques, curated based on several criteria.
The selected machine learning methods are well-established within the literature and share specific architectural traits. While their standard implementations operate in a non-iterative state without utilizing iterative greedy feature selection or pruning, they are inherently compatible with iterative frameworks. Unlike static filters such as Mutual Information \cite{kraskov2004estimating} or variance thresholding \cite{guyon2003introduction} where feature importance remains constant—these methods produce importance scores sensitive to the evolving feature space, sensibly allowing for iterative re-evaluation. Furthermore, to maintain invariance to distance concentration, we explicitly exclude methods reliant on distance calculations. This avoids the high-dimensional loss of contrast phenomenon, where the relative difference between nearest and farthest neighbor distances vanishes \cite{beyer1999nearest}, ensuring the selection process is not compromised by the instability of distance metrics in high-dimensional spaces.

In the following, we provide a formal description of the feature selection methods utilized in this study:

\subsubsection{Embedded Tree-Based Importance}
We utilize Tree-based estimators, specifically Random Forests \cite{breiman2001random} and XGBoost \cite{chen2016xgboost}. These two methods compute feature importance through Mean Decrease in Impurity (MDI). For a node $\tau$ and a feature $f$, the importance is calculated as the sum of weighted impurity decreases $p(\tau)\Delta i(\tau)$ across all trees in the ensemble.

\subsubsection{ReliefF}
ReliefF \cite{kononenko1994estimating} is a non-parametric filter method that evaluates features based on their ability to distinguish between neighboring instances. It updates a weight vector by comparing an instance with its $k$-nearest neighbors from the same class (hits) and different classes (misses).

\subsubsection{LASSO Regularization}
The Least Absolute Shrinkage and Selection Operator (LASSO) \cite{tibshirani1996regression} introduces an $L_1$ penalty to the loss function:
\begin{equation}
    \min_{\beta} \left( \| Y - X\beta \|^2 + \alpha \|\beta\|_1 \right)
\end{equation}
The sparsity-inducing nature of the $L_1$ norm performs feature selection by shrinking the coefficients of non-informative features to zero.

\subsubsection{Permutation Importance}
Permutation importance \cite{altmann2010permutation} is a model-agnostic approach that measures the drop in model performance when the values of a single feature are randomly shuffled. This breaks the association between the feature and the outcome, providing a direct measure of feature dependence.

\section{Methodology} \label{sec:meth}
Traditional feature ranking methodologies typically evaluate the importance of a feature $f_i$ based on the complete feature set $\mathcal{F}$. However, such static rankings often fail to account for feature multi-collinearity and complex interaction effects. In scenarios where redundant features are present, the individual contribution of a variable may be artificially diluted or masked. If high dimensionality obscures the data and notions like distance \cite{beyer1999nearest}, it is also possible that the feature selection process might be effected by the excessive number of dimensions.

To address this, we implement an iterative greedy strategy, conceptually grounded in Recursive Feature Elimination. By iteratively removing the least significant feature and re-evaluating the feature importance scores, we ensure that the importance scores are consistently computed relative to the evolving feature manifold. This process continues until the search space is exhausted, providing a robust ranking that captures the persistence of a feature's predictive power.

Let $\mathcal{A}$ be a feature importance estimator (e.g., Random Forest, XGBoost, or LASSO) that maps a dataset $(\mathbf{X}, \mathbf{y})$ (or $(\mathbf{X})$ if unsupervised) to a weight vector $\mathbf{w} \in \mathbb{R}^d$, where $d$ is the number of active features. We define the iterative pruning process as a sequence of subsets $S_0, S_1, \dots, S_n$, where $S_0 = \{1, \dots, n\}$. At each step $t$, the least important feature $f^*$ is identified and removed:

\begin{equation}
    f^* = \arg\min_{j \in S_t} \{ \mathcal{A}(\mathbf{X}_{S_t}, \mathbf{y})_j \}
\end{equation}

The elimination order is stored in a sequence $\epsilon$, which eventually defines the final global ranking. Features that survive the pruning process longest (i.e., those appearing late in $\pi$) are considered the most significant. This approach is agnostic to the underlying base estimator, provided the estimator yields a reliable feature importance measurement.

\begin{figure}[h]
    \centering
    \includegraphics[width=\linewidth]{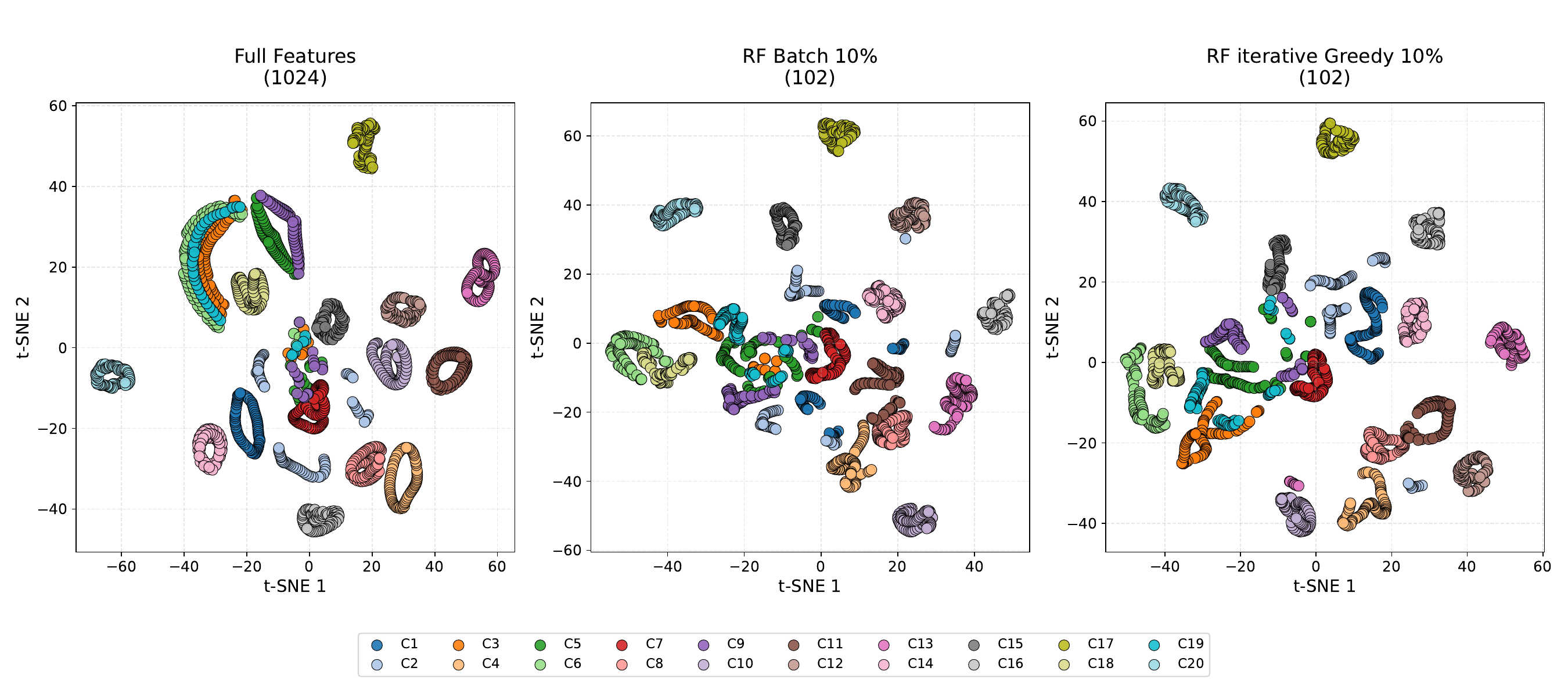}
    \caption{The projection of the COIL-20 dataset using t-SNE, presented for the full feature set, and the top 10\%, selected by both the normal and the iterative greedy approach.}
    \label{fg:tsne}
\end{figure}

In Fig.~\ref{fg:tsne}, we can see the projection of the COIL-20~\cite{Neneetal1996} dataset using t-SNE~\cite{VandermaatenHinton2008} with the full feature set, and the selection of the top 10\% of the features using both the normal and the iterative approach based on the Random Forest~\cite{breiman2001random}. It is evident that there is a discrepancy between these figures. The t-SNE projection for each case is different, with the iterative approach seemingly offering a better discrimination between the classes. This observation is in consensus with our assumptions: i) The curse of dimensionality obscures the feature selection process as well, with different results observed from the normal and the iterative approach. ii) The iterative approach seemingly yields better inter-class discrimination on COIL-20 dataset. This highlights the possibility that by gradually reducing the dimensionality and re-evaluating the feature importance values, we are likely to get better feature selection results. These observations and claims however, need to be verified using an extensive empirical evaluation.

\section{Experimental Setup}\label{sec:expset}
The empirical evaluation is conducted using the \texttt{FSEVAL} benchmarking suite \cite{rajabinasab2026fsevalfeatureselectionevaluation}, a specialized framework designed for the extensive and comprehensive evaluation of feature selection algorithms. The framework automates the life-cycle of the experiment, generating nested feature subsets based on the computed importance rankings. For each dataset, performance is evaluated across a granular scale ranging from $5\%$ to $100\%$ of the total feature count, with increments of $5\%$.  To ensure the generalizability of the results and mitigate the risk of overfitting, we employ a 5-fold cross-validation (CV) scheme. Within each fold, we calculate the \textit{Accuracy} (ACC) and \textit{Area Under the Curve} (AUC) for the selected subsets. Unsupervised measures such as \textit{Clustering Accuracy} (CLSACC) and \textit{Normalized Mutual Information} (NMI) are also calculated. \textit{Average Angle Difference} (AAD) \cite{rajabinasab2025metrics} is calculated as a model-agnostic evaluation metric. FSDEM \cite{rajabinasab2025fsdem} is also used on top of these measures to provide overall overview of the experimental results and facilitate the measurement of feature selection stability. We also provide a scalability analysis to show how much additional computational cost the iterative approach introduces.

\subsection{Algorithmic Configurations}
Five base estimators were selected to provide a diverse representation of feature selection strategies. Each estimator is configured to balance computational efficiency with robust importance estimation:
\begin{itemize}
    \item \textbf{Random Forest (RF) \& XGBoost:} Both ensemble methods utilize 50 estimators to ensure stable impurity-based (MDI) rankings.
    \item \textbf{ReliefF:} Configured with $k=10$ neighbors to evaluate the local manifold structure and capture feature dependencies.
    \item \textbf{LASSO:} Implemented via Logistic Regression with an $L_1$ penalty ($C=0.5$). To accommodate multiclass datasets, the model is wrapped in a \textit{One-Vs-Rest} (OvR) architecture.
    \item \textbf{Permutation Importance:} Evaluated using a Random Forest backbone with 20 estimators and 3 shuffle iterations per feature.
\end{itemize}

\subsection{Dataset Characteristics}
The benchmark is performed on a diverse collection of datasets from Penn Machine Learning Benchmarks \cite{Olson2017PMLB}. The characteristics of these datasets, including the number of instances, features, and classes, are summarized in Table \ref{tab:datasets}.

\begin{table}[h]
\centering
\caption{Summary of datasets used in this study.}
\label{tab:datasets}
\begin{tabular}{@{}lrrr@{}}
\hline
Dataset Name & Instances & Features & Classes \\
\hline
agaricus\_lepiota & 8145 & 22 & 2 \\
allbp & 3772 & 29 & 3 \\
analcatdata\_authorship & 841 & 70 & 4 \\
ann\_thyroid & 7200 & 21 & 3 \\
backache & 180 & 32 & 2 \\
calendarDOW & 399 & 32 & 5 \\
churn & 5000 & 20 & 2 \\
clean1 & 476 & 168 & 2 \\
clean2 & 6598 & 168 & 2 \\
collins & 485 & 23 & 13 \\
dermatology & 366 & 34 & 6 \\
dna & 3186 & 180 & 3 \\
flags & 178 & 43 & 5 \\
hypothyroid & 3163 & 25 & 2 \\
ionosphere & 351 & 34 & 2 \\
led24 & 3200 & 24 & 10 \\
mfeat\_fourier & 2000 & 76 & 10 \\
mfeat\_pixel & 2000 & 240 & 10 \\
molecular\_biology\_promoters & 106 & 57 & 2 \\
movement\_libras & 360 & 90 & 15 \\
mushroom & 8124 & 22 & 2 \\
ring & 7400 & 20 & 2 \\
satimage & 6435 & 36 & 6 \\
soybean & 675 & 35 & 18 \\
spect & 267 & 22 & 2 \\
spectf & 349 & 44 & 2 \\
splice & 3188 & 60 & 3 \\
tokyo1 & 959 & 44 & 2 \\
\hline
\end{tabular}
\end{table}

The datasets are selected to be diverse, constituting a different number of instances, features, and classes. We keep the maximum dimensionality of the datasets included in this study relatively small (240). This is because that we believe going to very high-dimensional spaces, which in real-world datasets usually results in sparse datasets (very small number of instances compared to the dimensionality). We believe that sparse datasets are not suitable for this study. In these datasets, many different combinations of features can constitute the same level of effectiveness, which might make a granular comparison very difficult.

\section{Experimental Results} \label{sec:resu}
Our experimental design leverages a diverse benchmark consisting of 28 datasets of varying dimensionality and complexity, ensuring the generalizability of our findings. To provide a holistic view of performance, we utilize five distinct evaluation metrics: Accuracy (ACC), Area Under the ROC Curve (AUC), Class-Weighted Accuracy (CLSACC), Normalized Mutual Information (NMI), and Average Angle Difference (AAD). By comparing the iterative variants against their static counterparts across these metrics, we aim to quantify the impact of recursive feature elimination on the feature selection performance.

\subsection{Feature Ranking Similarity}

To study whether the iterative implementation of the feature selection algorithms identify the same features as important,  we analyze the \textit{Feature Selection Similarity}, which measures the similarity between the rankings produced by different selectors.

Let $M_1$ and $M_2$ be two feature selection methods. For a given selection budget $k$ features to be selected, let $S_1 \subseteq F$ and $S_2 \subseteq F$ represent the sets of the top-$k$ features selected by $M_1$ and $M_2$, respectively, where $F$ is the total set of features. The similarity $R_S$ at cardinality $k$ is defined as the normalized cardinality of the intersection:
\begin{equation}
    R_S(S_1, S_2, k) = \frac{|S_1 \cap S_2|}{k}
\end{equation}
where $R_S \in [0, 1]$. A value of 1 indicates identical feature preferences, while a value of 0 indicates completely different selections. In our analysis, we express $k$ as a percentage of the total feature count ($p \in \{5\%, 10\%, 15\%, 20\%, 25\%\}$) to observe how similar to their iterative implementation the feature selection algorithms perform.

\begin{figure}[h]
    \centering
    \includegraphics[width=0.7\linewidth]{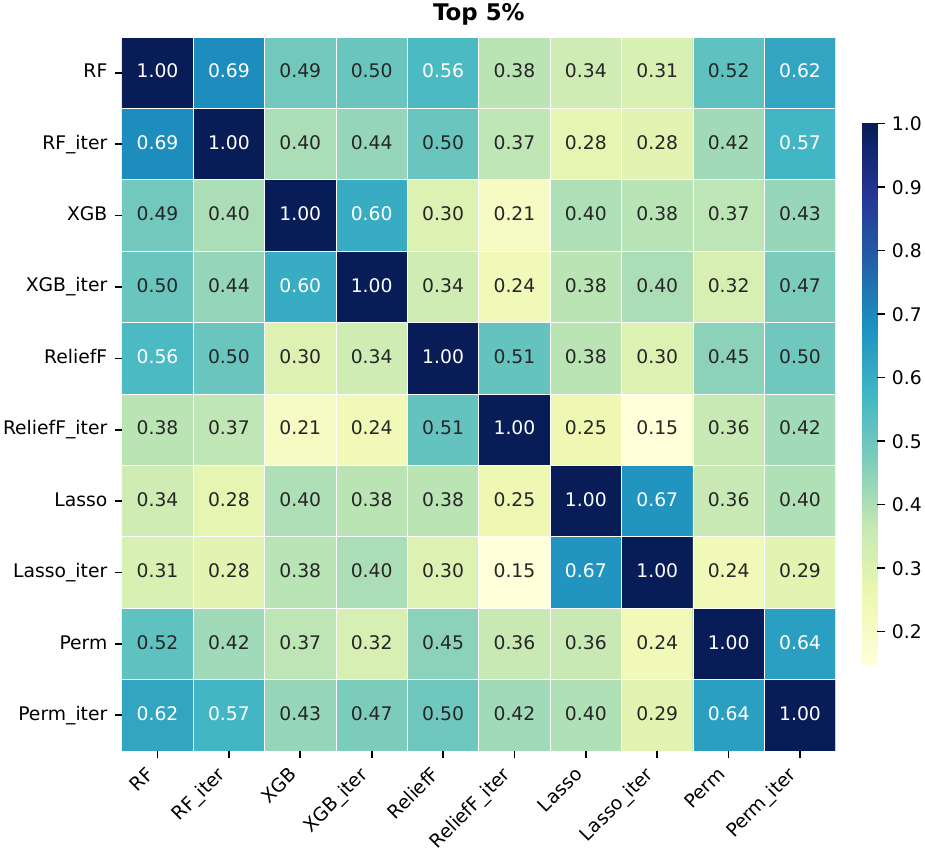}%
    \hfill
    \includegraphics[width=0.43\linewidth]{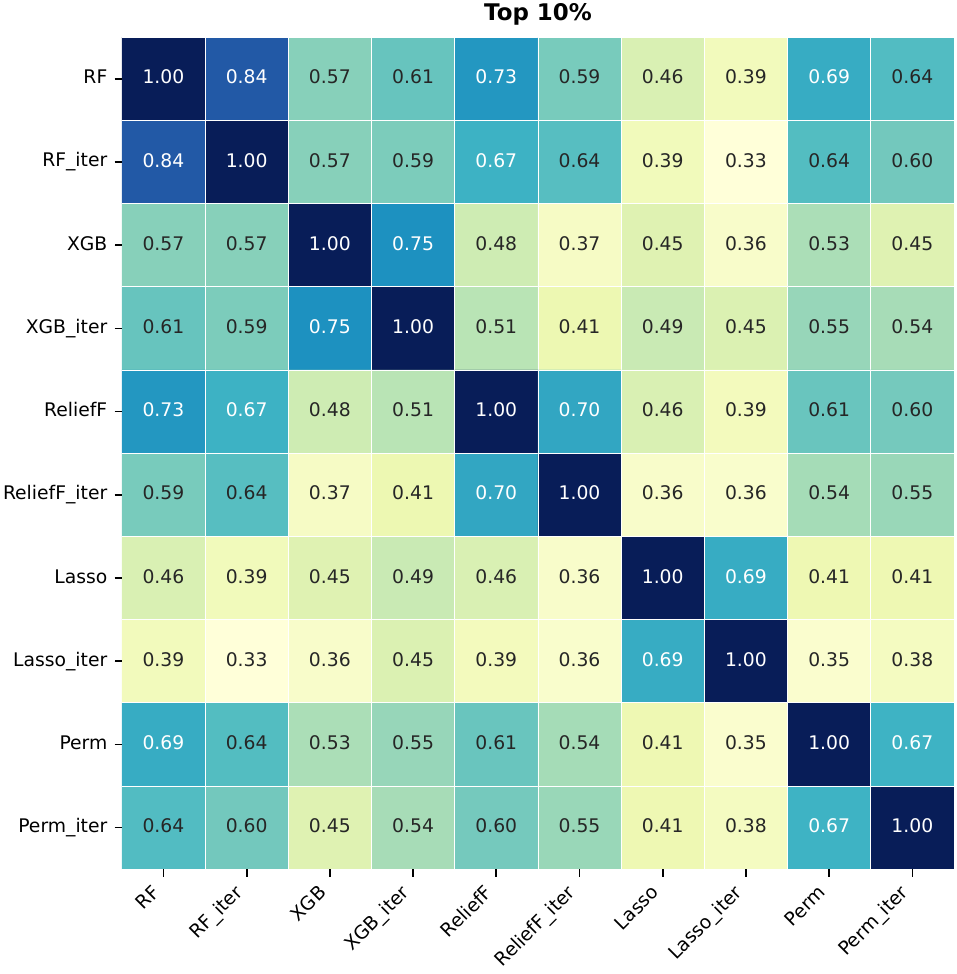}
    \includegraphics[width=0.43\linewidth]{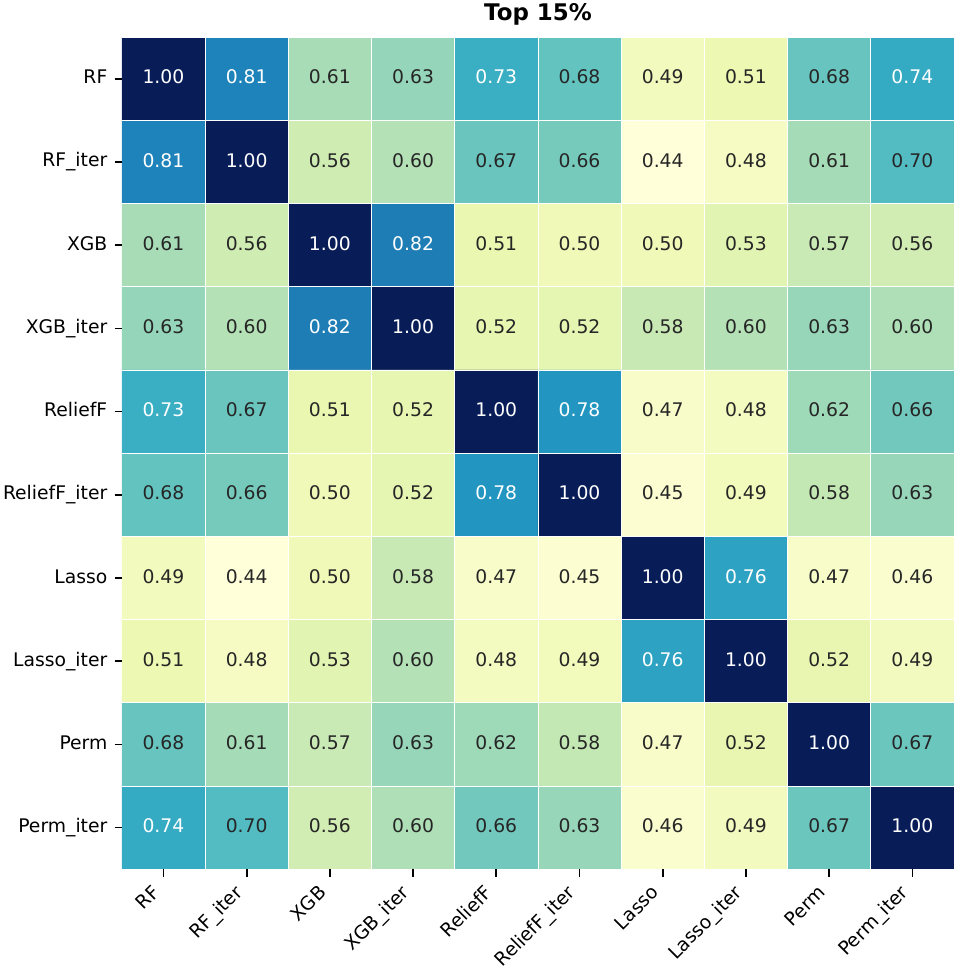}
    \includegraphics[width=0.43\linewidth]{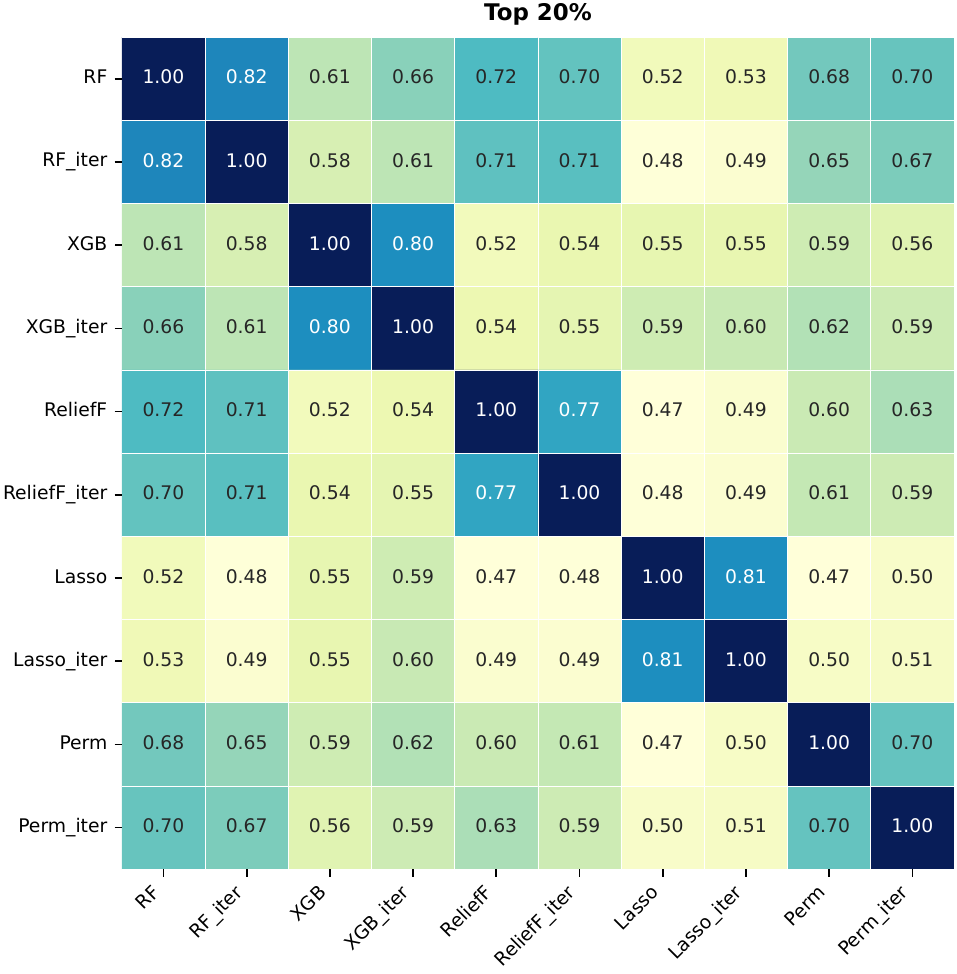}
    \includegraphics[width=0.43\linewidth]{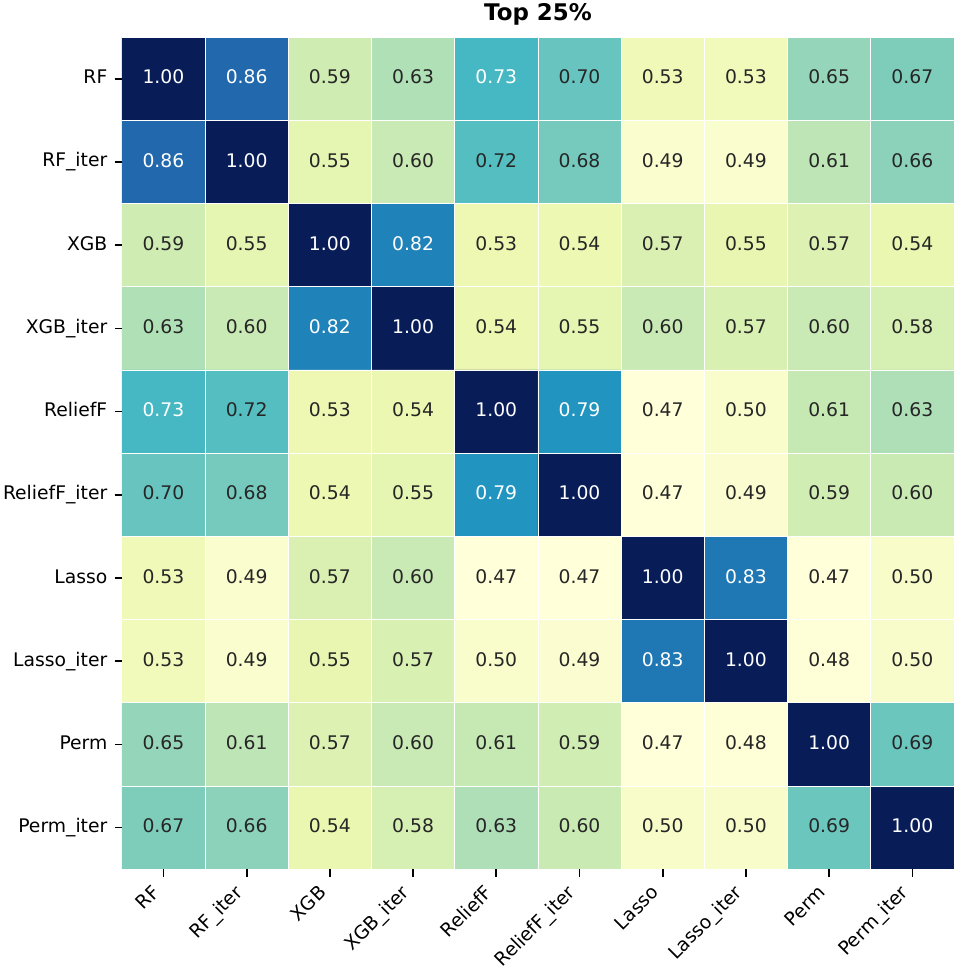}
    \caption{Feature Ranking Similarity heatmaps for selection budgets of 5\% (top), and 10\%--25\% (bottom row), reflecting a high amount of discrepancy between the normal approaches and their greedy counterpart.}
    \label{fg:heatmaps}
\end{figure}

Figure~\ref{fg:heatmaps} illustrates the mean overlap across all benchmark datasets for various selection budgets.
The heatmap clearly shows that there is a significant difference in the features selected by a feature selection method compared to its iterative implementation. This difference becomes lower as more features are selected, and it is higher in top ranking features set. This motivates a more in-depth study on both, two show if there is a performance difference, and if there is, which one is performing better.

\subsection{Performance Analysis}
Predictive performance is evaluated through five distinct lenses, ranging from classification performance to clustering and model-agnostic evaluation. This multi-faceted approach allows us to determine if the iterative implementation of the feature selection methods consistently improves the quality of the selected features across diverse data topologies.

\subsubsection{Analysis of Accuracy (ACC)}
\begin{figure}[h]
    \centering
    \includegraphics[width=\linewidth]{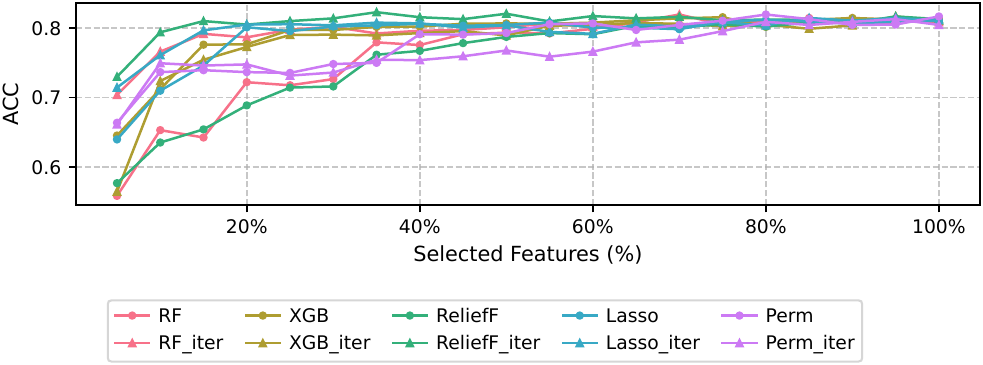}
    
    \includegraphics[width=0.8\textwidth]{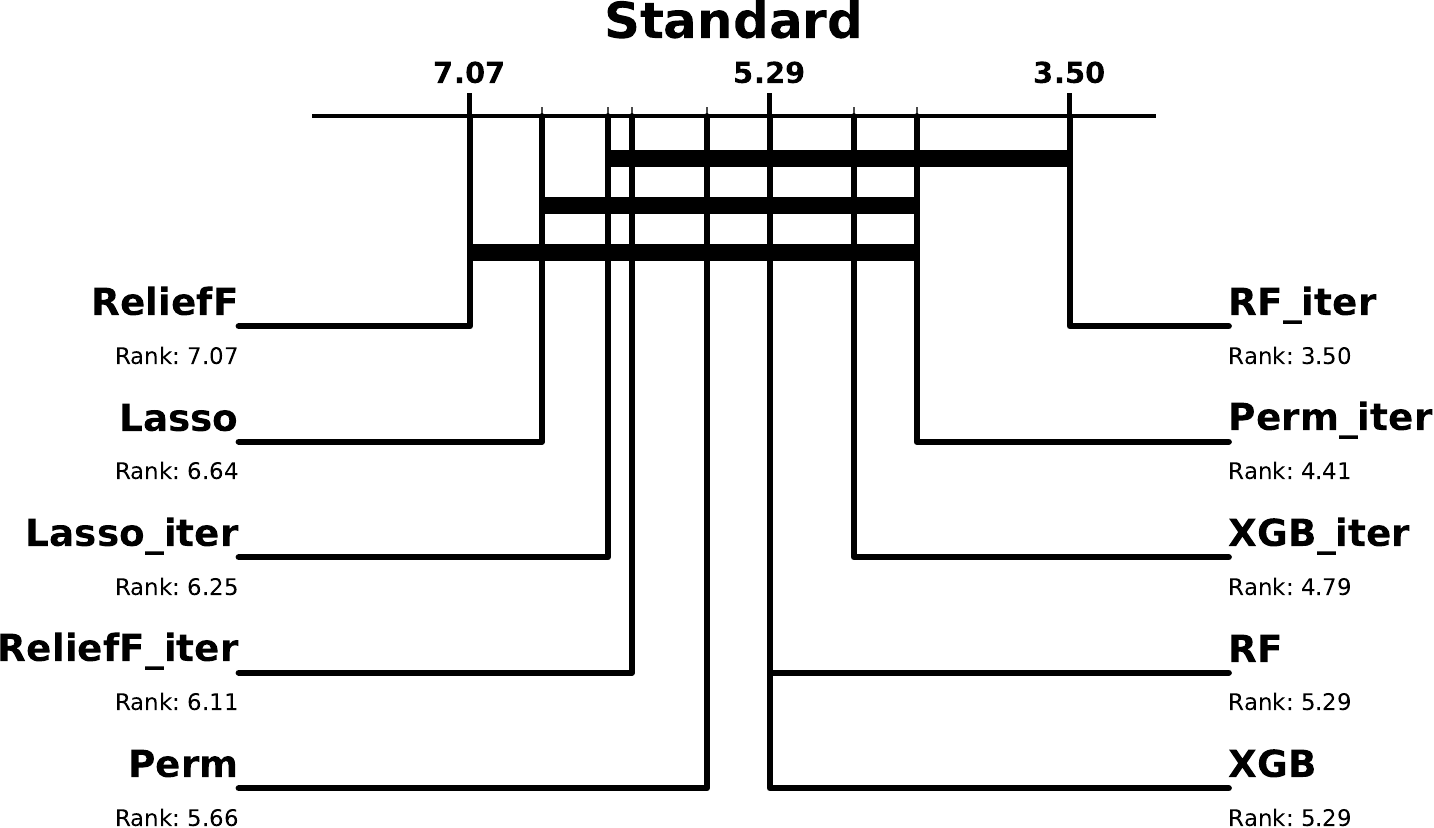}
    \includegraphics[width=0.8\textwidth]{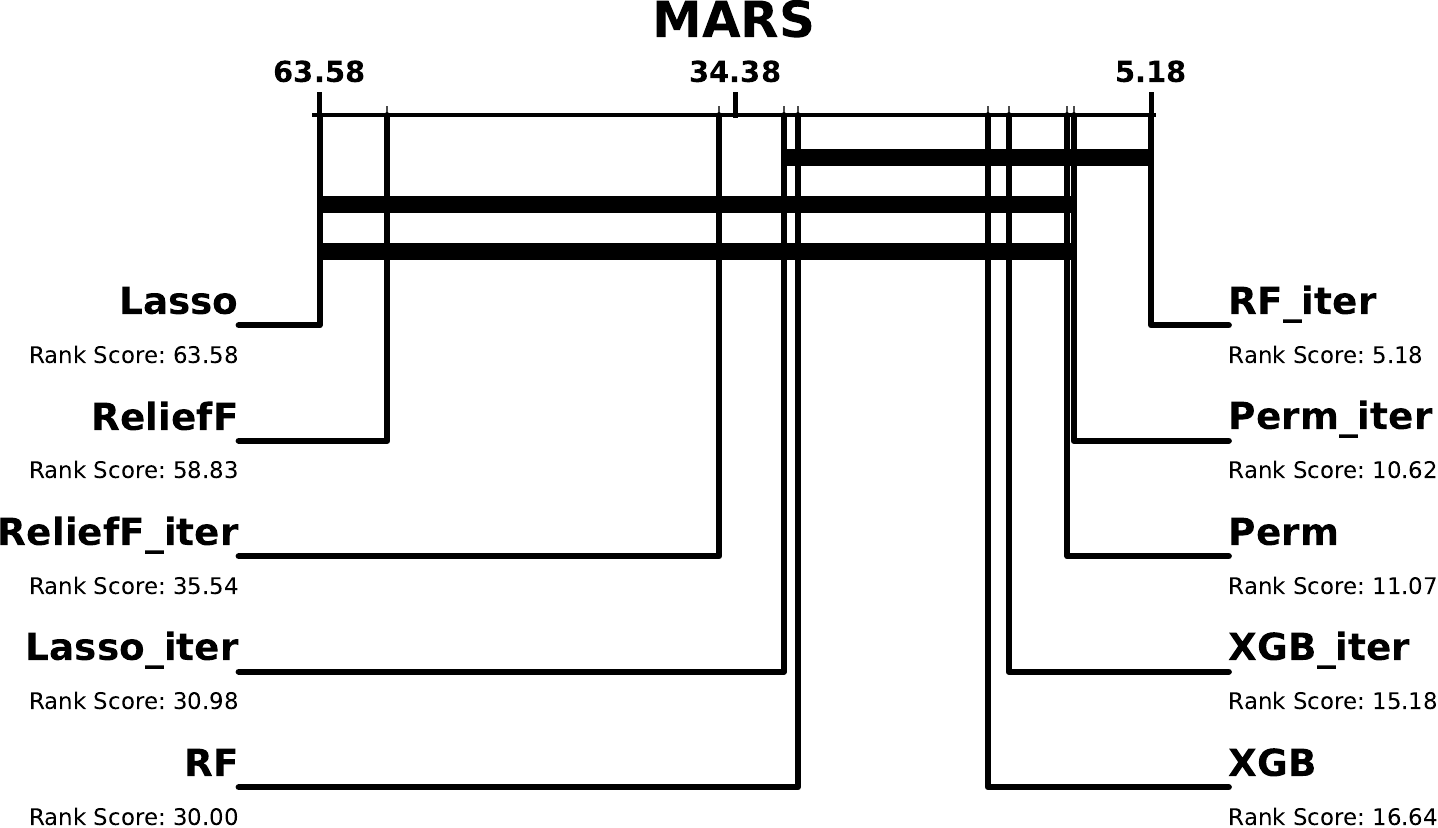}
    \caption{Comparative performance based on Accuracy (ACC). The  line plot tracks the \textit{movement\_libras} dataset as a case study, followed by a Critical Difference diagrams over all datasets and average performance of the methods. Both MARS and Standard rank statistics agree on the superiority of the iterative approach. Only minor changes of positions observed with the inclusion of metric values in MARS.}
    \label{fg:acc}
\end{figure}

Accuracy serves as the primary indicator of overall classification success. Alongside with AUC, it is the most important evaluation metric we use, as the feature selection methods we selected for this study are all supervised. In this analysis, we study how different sizes of selected feature subsets perform based on different feature selection methods compared to their iterative counterpart. We present the performance plot of the \textit{movement\_libras} dataset as an example of the extensive set of datasets we use. We also provide the rank analysis based on the standard rank \cite{Demsar2006} and Magnitude-Aware Rank Statistics (MARS)\cite{MARS} and the corresponding critical difference diagrams over the average performance of the methods on all datasets. The performance plot and the Critical Difference diagrams based on ACC are presented in Fig.~\ref{fg:acc}. 

We observe that, almost consistently, the iterative approach works better than the standard one, confirming our expectations about the effect of high-dimensionality in degrading the feature selection performance. The Critical Difference diagrams show that the iterative approaches consistently improve the feature selection performance and provide superior results compared to the corresponding standard implementation.

\subsubsection{Analysis of Area Under the ROC Curve (AUC)}
\begin{figure}[h]
    \centering
    \includegraphics[width=\linewidth]{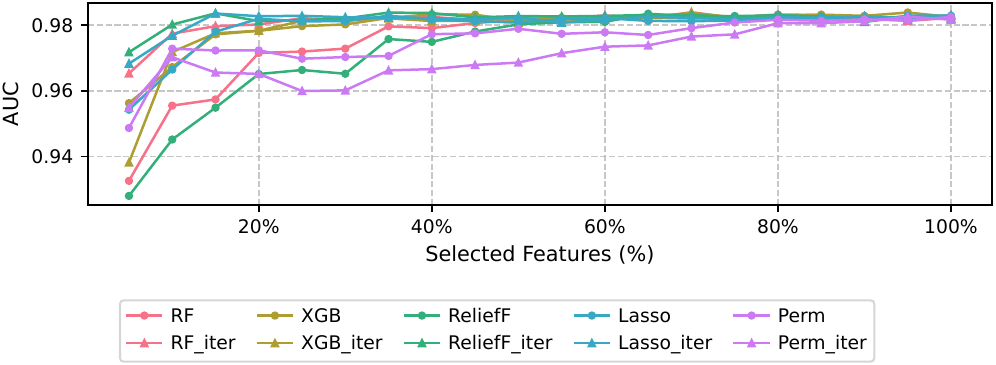}
    
    \includegraphics[width=0.8\textwidth]{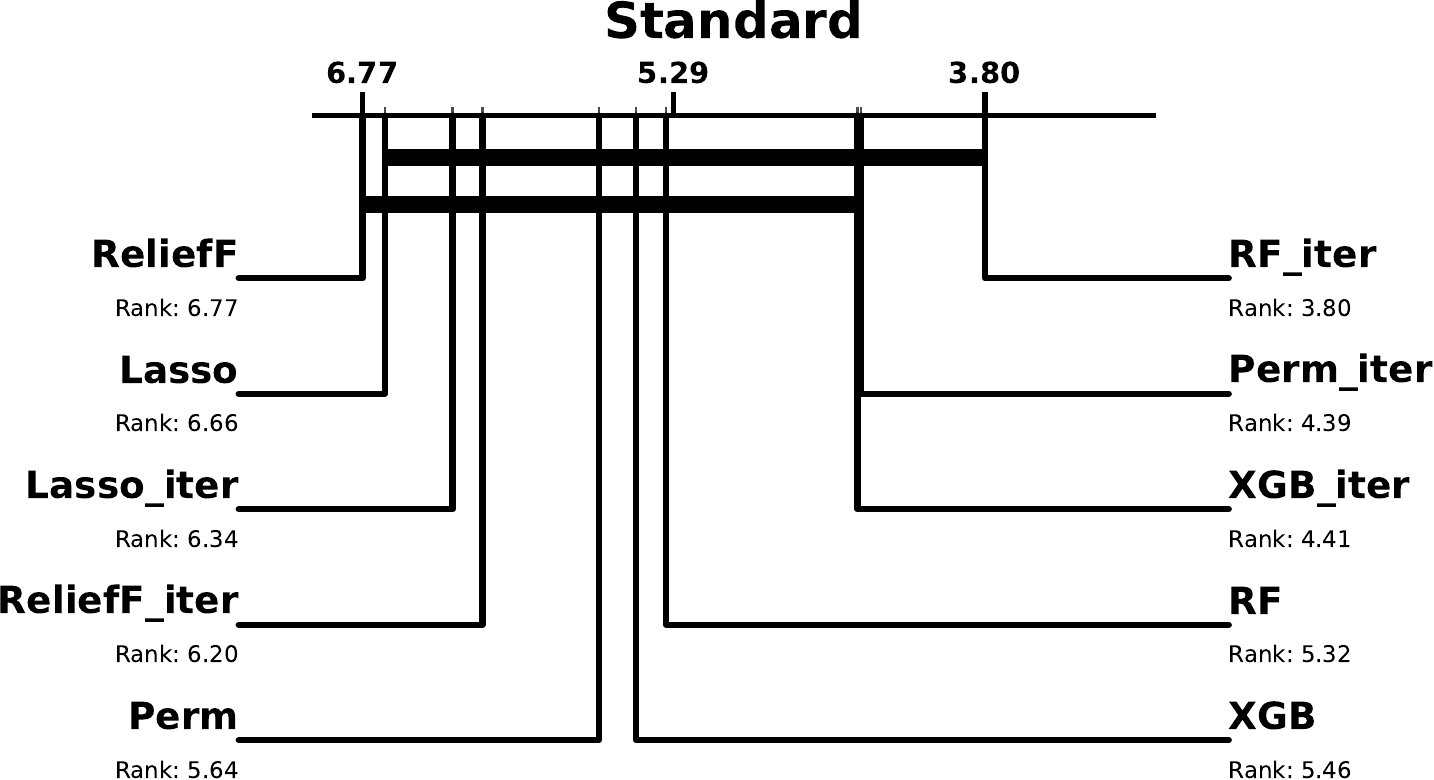}
    \includegraphics[width=0.8\textwidth]{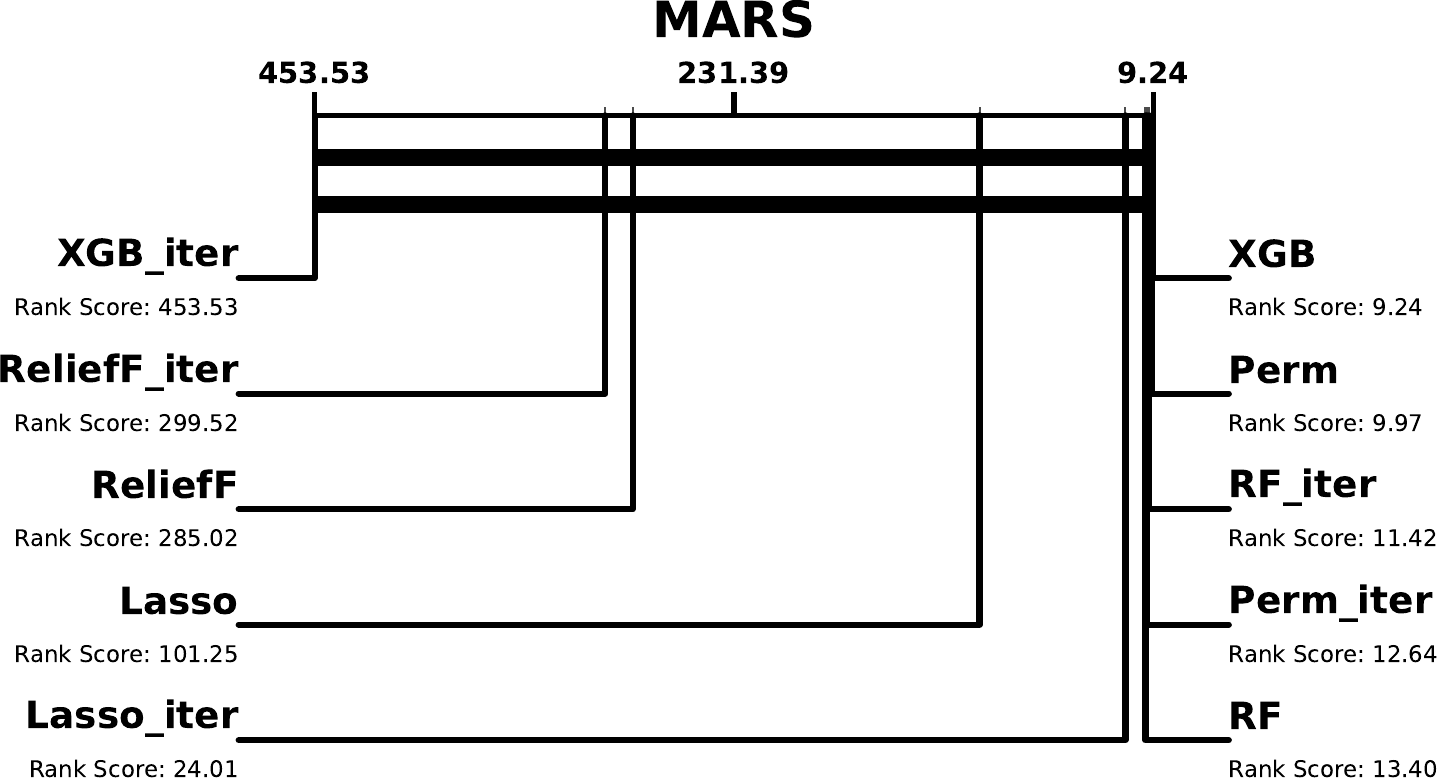}
    \caption{Comparative performance based on AUC. The  line plot tracks the \textit{movement\_libras} dataset as a case study, followed by Critical Difference diagrams over all datasets and average performance of the methods. Both MARS and Standard rank statistics agree on the superiority of the iterative approach. Only minor changes of positions observed with the inclusion of metric values in MARS.}
    \label{fg:auc}
\end{figure}

The Area Under the ROC Curve (AUC) serves as a threshold-independent measure of the model's discriminative power. Given that our study focuses on supervised feature selection, AUC is also considered  as a cornerstone metric for evaluating the class separability of the chosen subsets. We analyze how different selection budgets impact this discriminative ability, comparing standard implementations against our proposed iterative framework. Fig.~\ref{fg:auc} presents the methods' behavior on the \textit{movement\_libras} dataset alongside the rank analyses across all 28 datasets.

The results indicate that the iterative approach almost consistently outperforms the static one, validating our hypothesis that this approach mitigates the performance loss typically caused by high-dimensional noise. As shown in the Critical Difference diagrams, the iterative variants occupy superior rank positions, demonstrating that they provide more reliable and robust feature rankings compared to the standard one-pass selection methods.

\subsubsection{Analysis of Clustering Accuracy (CLSACC)}
\begin{figure}[h]
    \centering
    \includegraphics[width=\linewidth]{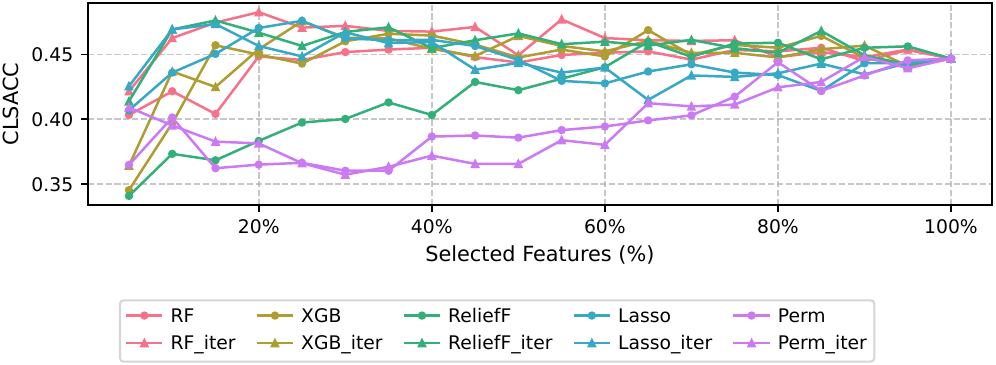}
    
    \includegraphics[width=0.8\textwidth]{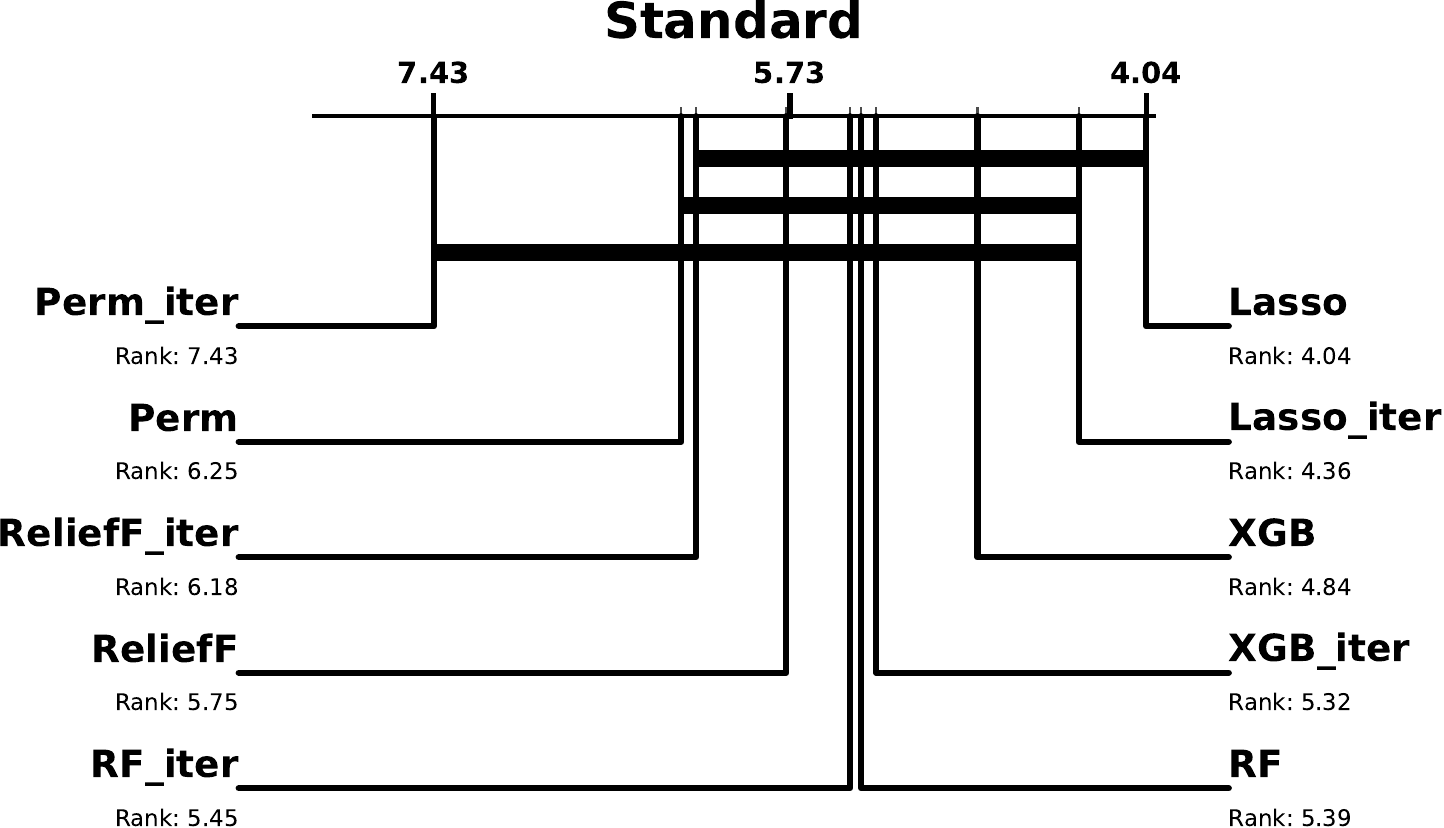}
    \includegraphics[width=0.8\textwidth]{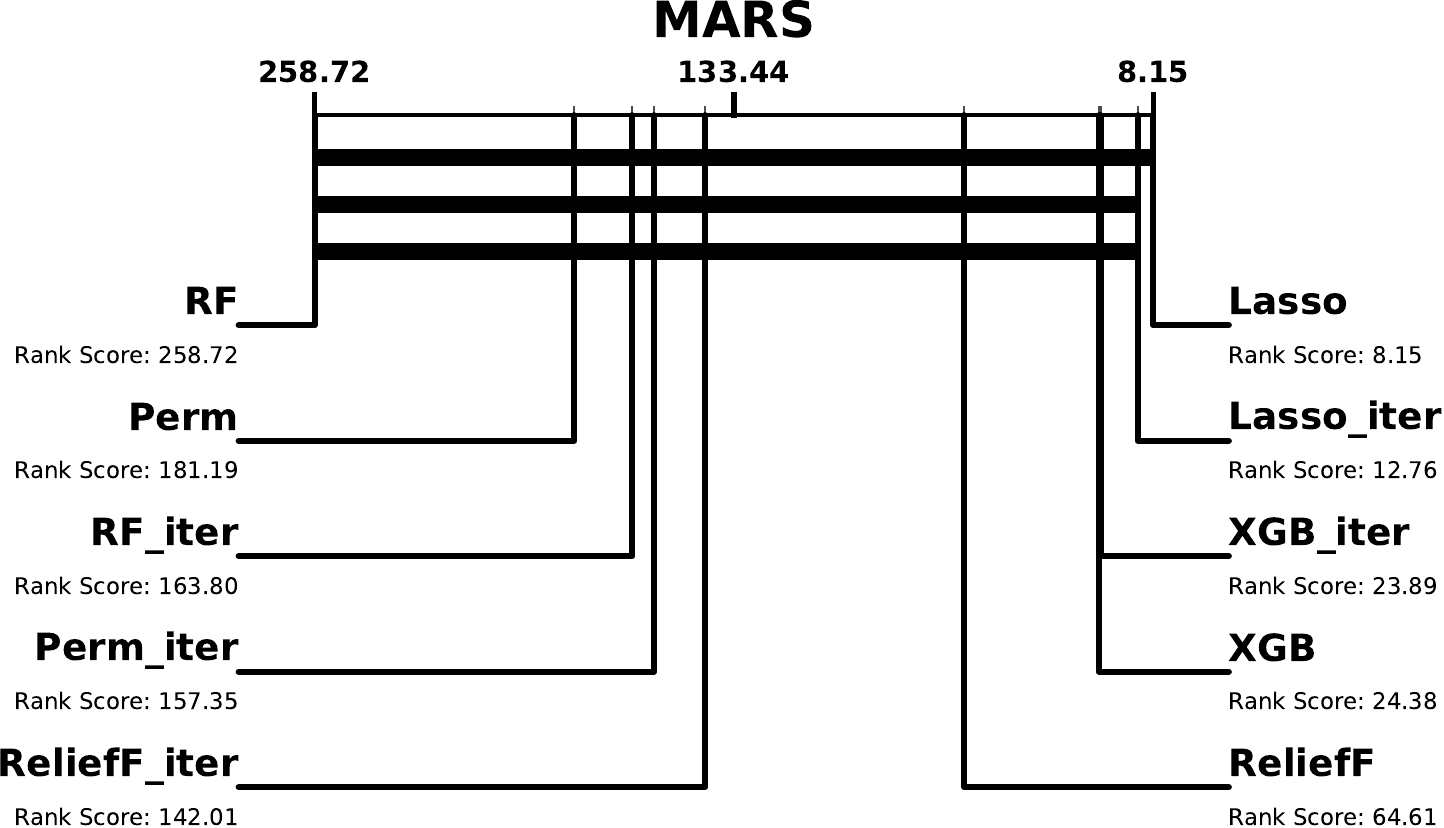}
    \caption{Comparative performance based on Clustering Accuracy (CLSACC). The line plot tracks the \textit{movement\_libras} dataset as a case study, followed by Critical Difference diagrams over all datasets and average performance of the methods. Both MARS and Standard rank statistics agree on the superiority of the iterative approach. MARS emphasizes on the superiority of metric values in iterative implementations.}
    \label{fg:clsacc}
\end{figure}

To also include the unsupervised downstream tasks, we also present the performance based on the Clustering Accuracy. This metric aims to map the clustering results labels to ground-truth labels, and hence, is very unstable compared to the supervised metrics. Fig.~\ref{fg:clsacc} presents the methods behavior on the \textit{movement\_libras} dataset alongside the rank analyses across all 28 datasets.
Evidently, the performance of iterative implementation is not as consistently superior as for the supervised measure.

\subsubsection{Analysis of Normalized Mutual Information (NMI)}
We also report unsupervised performance using the Normalized Mutual Information (NMI). Unlike clustering accuracy, NMI evaluates the agreement between predicted clusters and ground-truth labels without requiring an explicit label mapping, making it more stable and less sensitive to permutation effects. 

\begin{figure}[h]
    \centering
    \includegraphics[width=\linewidth]{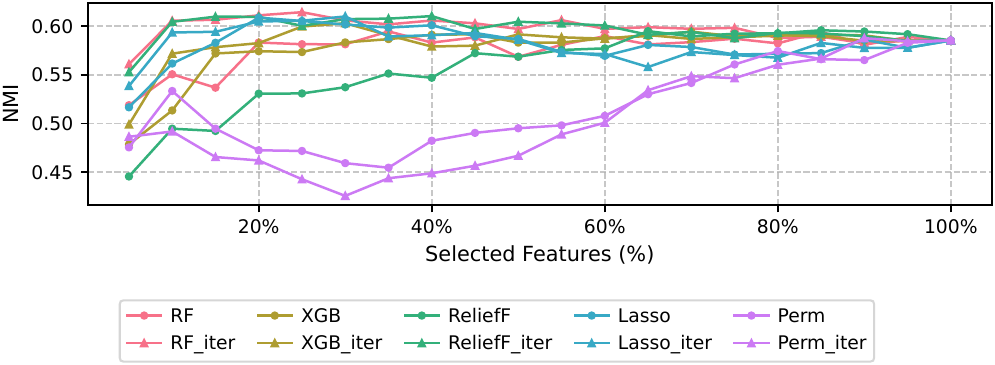}
    
    \includegraphics[width=0.8\textwidth]{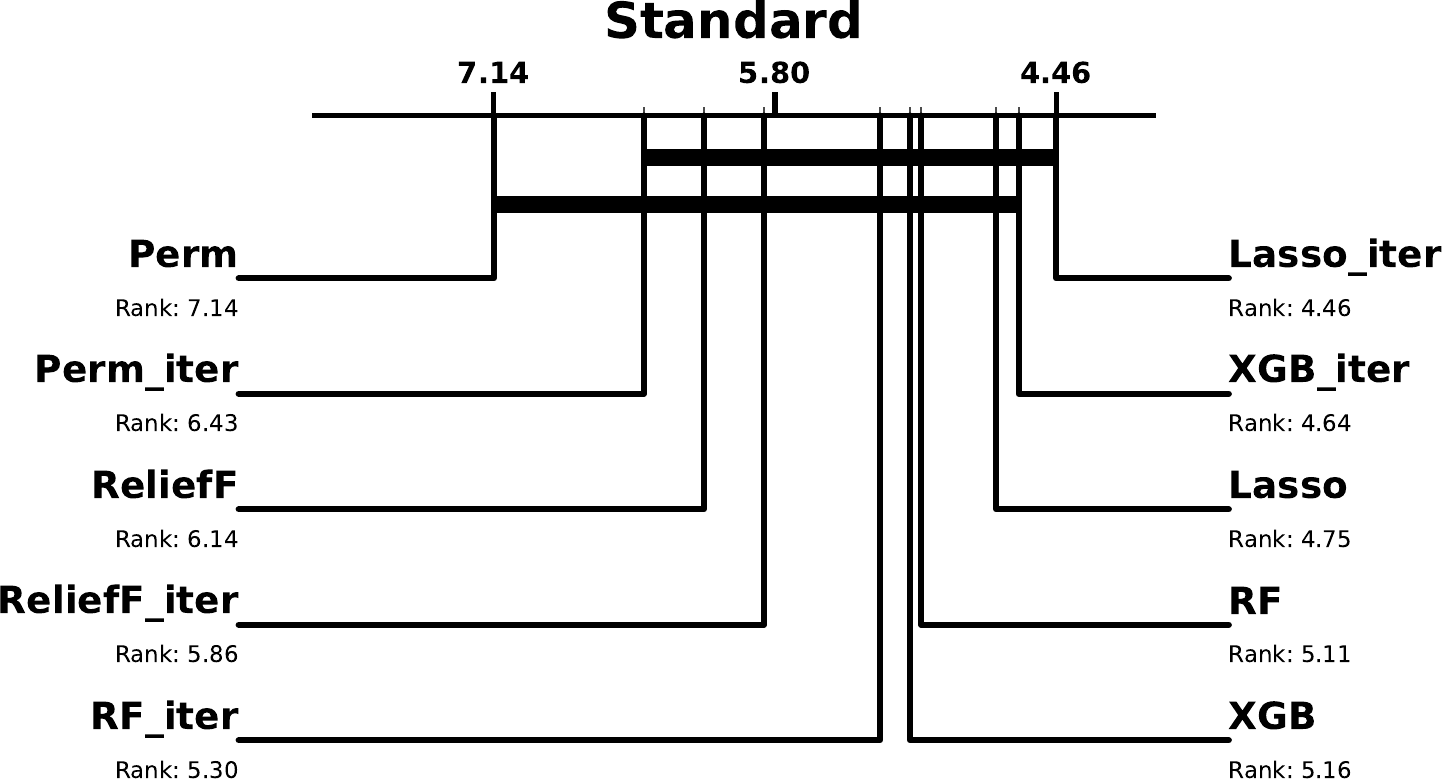}
    \includegraphics[width=0.8\textwidth]{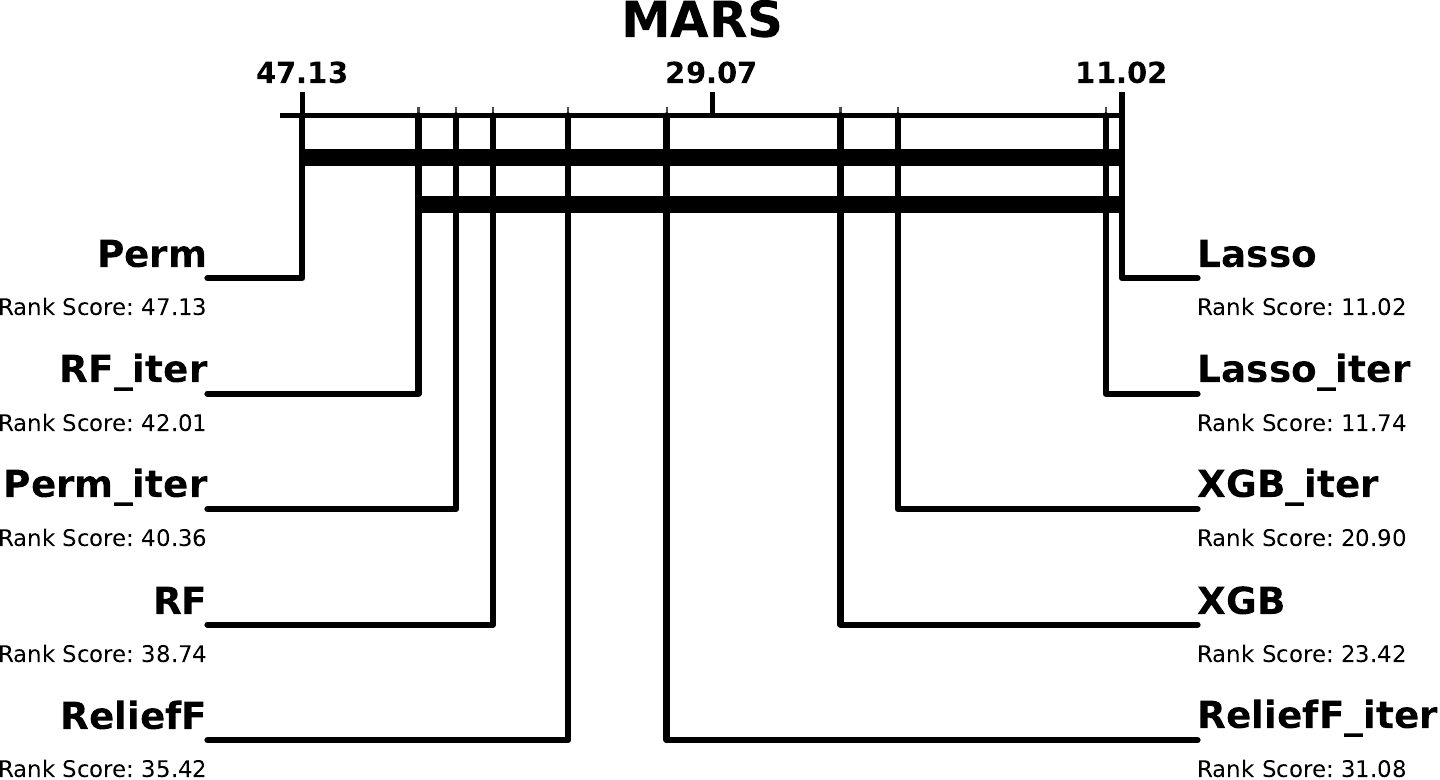}
    \caption{Comparative performance based on Normalized Mutual Information (NMI). The line plot tracks the \textit{movement\_libras} dataset as a case study, followed by Critical Difference diagrams over all datasets and average performance of the methods. Both MARS and Standard rank statistics agree on the superiority of the iterative approach. Only minor changes observed with the inclusion of metric values in MARS.}
    \label{fg:nmi}
\end{figure}

Fig.~\ref{fg:nmi} shows the behavior of the methods on the \textit{movement\_libras} dataset together with the global rank analysis across all 28 datasets. As expected, the iterative implementation exhibits greater variability under this unsupervised criterion compared to the supervised metrics. Nevertheless, it shows an almost consistent improvement over the standard implementation, highlighting the effectiveness of this approach, also for unsupervised downstream tasks.

\subsubsection{Analysis of Average Angle Difference (AAD)}
\begin{figure}[h]
    \centering
    \includegraphics[width=\linewidth]{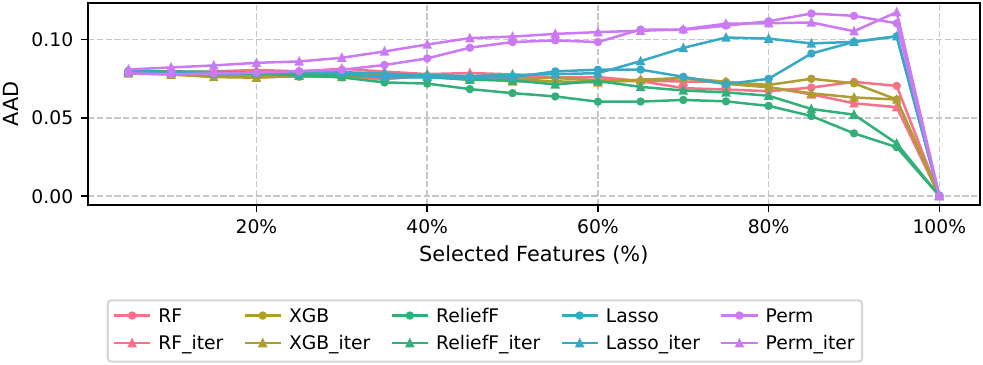}
    
    \includegraphics[width=0.8\textwidth]{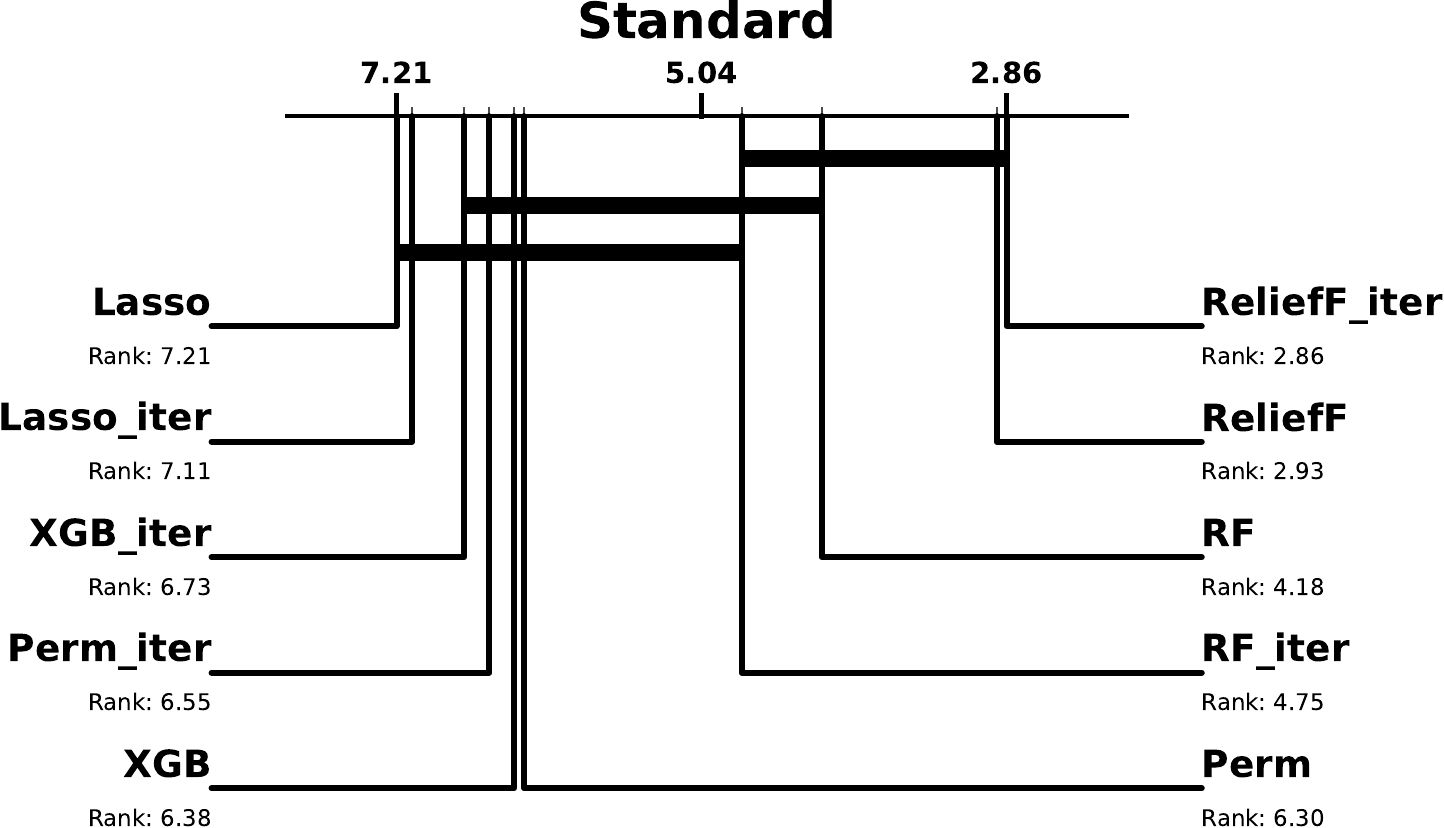}
    \includegraphics[width=0.8\textwidth]{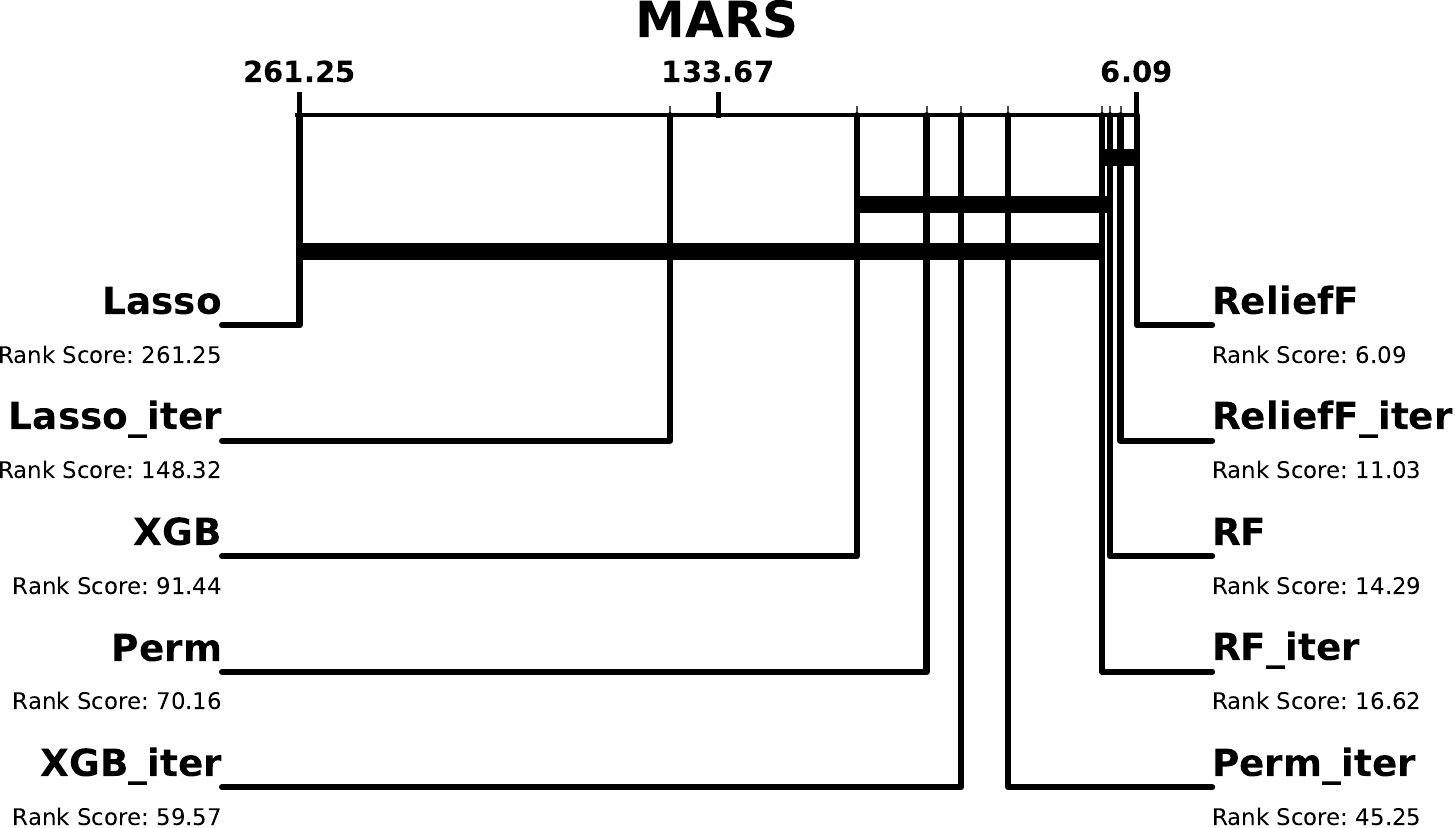}
    \caption{Comparative performance based on Average Angle Difference (AAD). The line plot tracks the \textit{movement\_libras} dataset as a case study, followed by Critical Difference diagrams over all datasets and average performance of the methods. Both MARS and Standard rank statistics agree on the superiority of the iterative approach. Only minor changes observed with the inclusion of AAD values in MARS.}
    \label{fg:aad}
\end{figure}

AAD \cite{rajabinasab2025metrics} quantifies the Average Angle Difference based on the first principal components, where lower values indicate better performance. It is a model-agnostic metric which favors the geometrical alignment based on Principal Component Analysis (PCA). In \textit{movement\_libras}, as shown in Fig.~\ref{fg:aad}, iterative methods demonstrate a sharp decline in the AAD value, however, it is still not as consistent as for supervised metrics. These findings make sense, as the supervised feature selection methods are expected to favor the improvement based on a supervised task. Showing somewhat competitive performance based on AAD, while improving based on both supervised and unsupervised metrics, overall shows the effectiveness of the proposed iterative approach.

\subsection{Scalability Analysis}

While the iterative framework consistently yields superior predictive performance, it is inherently more time-consuming than static methods. For a dataset with dimensionality $d$, the iterative approach requires repeating the feature selection process $d-1$ times to produce a full ranking. However, it should be noted that as the dimensionality is reduced at each recursive step, the execution time for the base selector also decreases, partially offsetting the overhead of the repetitions.

Figure~\ref{fg:scalability_comparison} illustrates the scalability of the methods. Despite the increased computational requirement, as the feature selection is most of the times, a one-time step, it can be considered feasible for high-dimensional datasets if the algorithm's computational complexity is not too high.

\section{Conclusion, Limitations, and Future Works} \label{sec:conc}
This paper investigated whether high dimensionality degrades the performance of feature selection methods in a manner similar to its well-known adverse effects on other concepts, such as distance measures. To address this question, we conducted a comparative study across a diverse and extensive collection of datasets, feature selection algorithms, and evaluation metrics. Our empirical results demonstrate that iteratively removing the least important feature and recalculating feature importance values consistently leads to superior performance when compared to applying the same feature selection algorithm only once. This

%

\FloatBarrier
\noindent finding suggests that feature importance estimates can be sensitive to the presence of less informative or redundant features, and that progressively refining the feature space allows the algorithm to better identify truly relevant variables.

Our analysis indicates that the primary limitation of this iterative approach lies in the number of times the feature selection process must be repeated. This requirement can become computationally expensive, particularly for high-dimensional datasets and for feature selection algorithms with  substantial computational complexity. Therefore, the scalability of the method becomes the main challenge that must be addressed before the approach can be widely adopted.

Addressing this computational limitation should therefore be a central focus of future research. Several possible strategies may help mitigate this issue. One straightforward modification is to remove the $r$ least important and potentially redundant features at each iteration, instead of eliminating only a single feature. This adjustment would reduce the total number of repetitions required while preserving the iterative refinement principle. Another potential approach is to partition the feature space into $p$ random subspaces and perform feature selection independently within each subspace. The least important feature, or the $r$ least important features, can then be removed from each subspace simultaneously. Such a strategy could reduce the number of sequential runs of the algorithm while also lowering the dimensionality handled in each run, thereby improving computational efficiency.

The investigation of these extensions, including their effectiveness, stability, and potential trade-offs, is beyond the scope of the present study. We leave a systematic evaluation of these strategies, along with other possible optimizations, as an important direction for future work.
\noindent
\begin{center}
    \includegraphics[width=\linewidth]{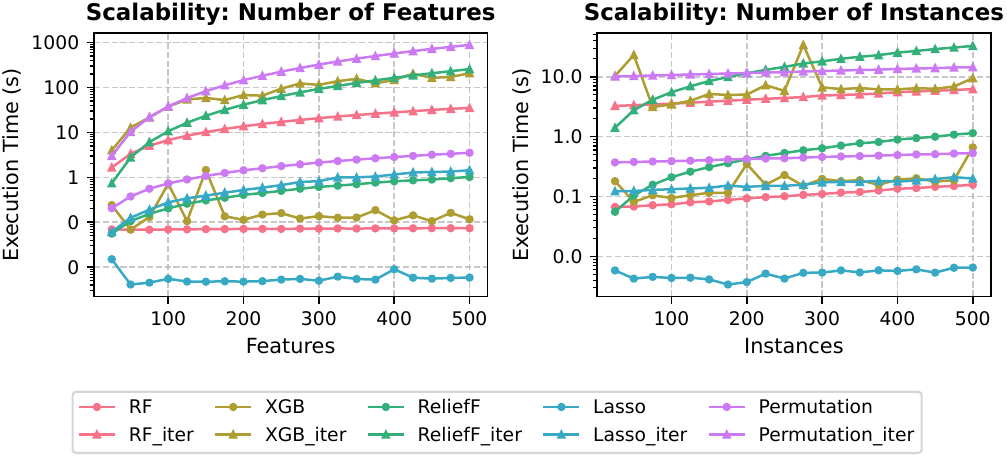}
    \captionof{figure}{Scalability of the standard and greedy approaches.}
    \label{fg:scalability_comparison}
\end{center}
\bibliographystyle{splncs04}
\bibliography{sisap2026}

@article{beyer1999nearest,
  title={When is “nearest neighbor” meaningful?},
  author={Beyer, Kevin and Goldstein, Jonathan and Ramakrishnan, Raghu and Shaft, Uri},
  booktitle={International conference on database theory},
  pages={217--235},
  year={1999},
  organization={Springer}
}

@article{guyon2003introduction,
  title={An introduction to variable and feature selection},
  author={Guyon, Isabelle and Elisseeff, Andr{\'e}},
  journal={Journal of machine learning research},
  volume={3},
  number={Mar},
  pages={1157--1182},
  year={2003}
}

@article{li2017feature,
  title={Feature selection: A data perspective},
  author={Li, Jundong and Cheng, Kewei and Wang, Suhang and Morstatter, Fred and Trevino, Robert P and Tang, Jiliang and Liu, Huan},
  journal={ACM computing surveys (CSUR)},
  volume={50},
  number={6},
  pages={1--45},
  year={2017},
  publisher={ACM New York, NY, USA}
}

@inproceedings{rajabinasab2025metrics,
  title={Metrics for Inter-Dataset Similarity with Example Applications in Synthetic Data and Feature Selection Evaluation},
    author = {Muhammad Rajabinasab and Anton Lautrup and Arthur Zimek},
    title = {Metrics for Inter-Dataset Similarity with Example Applications in Synthetic Data and Feature Selection Evaluation},
    booktitle = {Proceedings of the 2025 SIAM International Conference on Data Mining (SDM)},
    chapter = {},
    pages = {527-537},
    doi = {10.1137/1.9781611978520.57}
}

@inproceedings{rajabinasab2025fsdem,
  author       = {Muhammad Rajabinasab and
                  Anton Danholt Lautrup and
                  Tobias Hyrup and
                  Arthur Zimek},
  editorX       = {Edgar Ch{\'{a}}vez and
                  Benjamin B. Kimia and
                  Jakub Lokoc and
                  Marco Patella and
                  Jan Sedmidubsk{\'{y}}},
  title        = {A Dynamic Evaluation Metric for Feature Selection},
  booktitle    = {Similarity Search and Applications - 17th International Conference,
                  {SISAP} 2024, Providence, RI, USA, November 4-6, 2024, Proceedings},
  series       = {Lecture Notes in Computer Science},
  volume       = {15268},
  pages        = {65--72},
  publisher    = {Springer},
  year         = {2024},
  url          = {https://doi.org/10.1007/978-3-031-75823-2\_6},
  doi          = {10.1007/978-3-031-75823-2\_6},
  bibsource    = {dblp computer science bibliography, https://dblp.org}
}

@article{tibshirani1996regression,
  title={Regression shrinkage and selection via the lasso},
  author={Tibshirani, Robert},
  journal={Journal of the Royal Statistical Society: Series B (Methodological)},
  volume={58},
  number={1},
  pages={267--288},
  year={1996},
  publisher={Wiley Online Library}
}

@article{breiman2001random,
  title={Random forests},
  author={Breiman, Leo},
  journal={Machine learning},
  volume={45},
  pages={5--32},
  year={2001},
  publisher={Springer}
}

@inproceedings{chen2016xgboost,
  title={Xgboost: A scalable tree boosting system},
  author={Chen, Tianqi and Guestrin, Carlos},
  booktitle={Proceedings of the 22nd acm sigkdd international conference on knowledge discovery and data mining},
  pages={785--794},
  year={2016}
}

@inproceedings{kononenko1994estimating,
  title={Estimating attributes: Analysis and extensions of RELIEF},
  author={Kononenko, Igor},
  booktitle={European conference on machine learning},
  pages={171--182},
  year={1994},
  organization={Springer}
}

@article{guyon2002gene,
  title={Gene selection for cancer classification using support vector machines},
  author={Guyon, Isabelle and Weston, Jason and Barnhill, Stephen and Vapnik, Vladimir},
  journal={Machine learning},
  volume={46},
  pages={389--422},
  year={2002},
  publisher={Springer}
}

@article{asadi2021face,
  title={Face recognition using color and edge orientation difference histogram},
  author={Asadi Amiri, S and Rajabinasab, Muhammad},
  journal={Journal of AI and Data Mining},
  volume={9},
  number={1},
  pages={31--38},
  year={2021},
  doi={10.22044/jadm.2020.9376.2072},
  publisher={Journal of AI and Data Mining}
}

@article{RAJABINASAB2025126254,
title = {Randomized PCA forest for approximate k-nearest neighbor search},
journal = {Expert Systems with Applications},
volume = {281},
pages = {126254},
year = {2025},
issn = {0957-4174},
doi = {https://doi.org/10.1016/j.eswa.2024.126254},
xurl = {https://www.sciencedirect.com/science/article/pii/S095741742403121X},
author = {Muhammad Rajabinasab and Farhad Pakdaman and Arthur Zimek and Moncef Gabbouj}
}

@article{altmann2010permutation,
  title={Permutation importance: a corrected feature importance measure},
  author={Altmann, Andr{\'e} and Tolo{\c{s}}i, Laura and Sander, Oliver and Lengauer, Thomas},
  journal={Bioinformatics},
  volume={26},
  number={10},
  pages={1340--1347},
  year={2010},
  publisher={Oxford University Press}
}

@article{kraskov2004estimating,
  title={Estimating mutual information},
  author={Kraskov, Alexander and St{\"o}gbauer, Harald and Grassberger, Peter},
  journal={Physical review E},
  volume={69},
  number={6},
  pages={066138},
  year={2004},
  publisher={APS}
}

@article{Olson2017PMLB,
    author="Olson, Randal S. and La Cava, William and Orzechowski, Patryk and Urbanowicz, Ryan J. and Moore, Jason H.",
    title="PMLB: a large benchmark suite for machine learning evaluation and comparison",
    journal="BioData Mining",
    year="2017",
    month="Dec",
    day="11",
    volume="10",
    number="36",
    pages="1--13",
    issn="1756-0381"
}

@article{Demsar2006,
  author  = {Janez Dem\v{s}ar},
  title   = {Statistical Comparisons of Classifiers over Multiple Data Sets},
  journal = {JMLR},
  volume  = {7},
  pages   = {1--30},
  year    = {2006},
  doix     = {10.5555/1248547.1248548}
}

@article{rajabinasab2026fsevalfeatureselectionevaluation,
  author       = {Muhammad Rajabinasab and
                  Arthur Zimek},
  title        = {{FSEVAL:} Feature Selection Evaluation Toolbox and Dashboard},
  journal      = {CoRR},
  volume       = {abs/2604.18227},
  year         = {2026}
}

@article{MARS,
  author       = {Muhammad Rajabinasab and
                  Afsaneh M. Nejad and
                  Arthur Zimek},
  title        = {{MARS:} Magnitude-Aware Rank Statistics},
  journal      = {CoRR},
  volume       = {abs/2605.23563},
  year         = {2026} 
}

@techreport{Neneetal1996,
  author      = {Nene, S. A. and Nayar, S. K. and Murase, H.},
  title       = {Columbia Object Image Library (COIL-20)},
  institution = {Columbia University},
  year        = {1996},
  number      = {CUCS-005-96}
}

@article{VandermaatenHinton2008,
  author  = {van der Maaten, Laurens and Hinton, Geoffrey},
  title   = {Visualizing Data using t-SNE},
  journal = {Journal of Machine Learning Research},
  year    = {2008},
  volume  = {9},
  number  = {86},
  pages   = {2579--2605},
  url     = {https://www.jmlr.org/papers/v9/vandermaaten08a.html}
}

@article{DBLP:journals/tkde/FrancoisWV07,
  author       = {Damien Fran{\c{c}}ois and
                  Vincent Wertz and
                  Michel Verleysen},
  title        = {The Concentration of Fractional Distances},
  journal      = {{IEEE} Trans. Knowl. Data Eng.},
  volume       = {19},
  number       = {7},
  pages        = {873--886},
  year         = {2007}
}

@inproceedings{DBLP:conf/sisap/OkkelsTAZS25,
  author       = {Camilla Birch Okkels and
                  Erik Thordsen and
                  Martin Aum{\"{u}}ller and
                  Arthur Zimek and
                  Erich Schubert},
  title        = {Approximate Single-Linkage Clustering Using Graph-Based Indexes: MST-Based
                  Approaches and Incremental Searchers},
  booktitle    = {{SISAP}},
  series       = {Lecture Notes in Computer Science},
  pages        = {233--247},
  xpublisher    = {Springer},
  xyear         = {2025}
}

@inproceedings{DBLP:conf/sdm/Anderberg0CHMRZ24,
  author       = {Alastair Anderberg and
                  James Bailey and
                  Ricardo J. G. B. Campello and
                  Michael E. Houle and
                  Henrique O. Marques and
                  Milos Radovanovic and
                  Arthur Zimek},
  title        = {Dimensionality-Aware Outlier Detection},
  booktitle    = {{SDM}},
  pages        = {652--660},
  publisher    = {{SIAM}},
  year         = {2024}
}

@article{DBLP:journals/tifs/AmsalegBBEFHRN21,
  author       = {Laurent Amsaleg and
                  James Bailey and
                  Am{\'{e}}lie Barbe and
                  Sarah M. Erfani and
                  Teddy Furon and
                  Michael E. Houle and
                  Milos Radovanovic and
                  Xuan Vinh Nguyen},
  title        = {High Intrinsic Dimensionality Facilitates Adversarial Attack: Theoretical
                  Evidence},
  journal      = {{IEEE} Trans. Inf. Forensics Secur.},
  volume       = {16},
  xpages        = {854--865},
  year         = {2021}
}

@InProceedings{rajabinasab2025hdsvdd,
author="Rajabinasab, Muhammad
and Lautrup, Anton D.
and Schneider-Kamp, Peter
and Zimek, Arthur",
title="Towards Semi-supervised Subspace Learning for Outlier Detection in Big Data",
booktitle="Similarity Search and Applications",
year="2026",
publisher="Springer Nature Switzerland",
address="Cham",
pages="330--344",
isbn="978-3-032-06069-3"
}

@article{rajabinasab2026randomizedpcaforestunsupervised,
  author       = {Muhammad Rajabinasab and
                  Farhad Pakdaman and
                  Moncef Gabbouj and
                  Peter Schneider{-}Kamp and
                  Arthur Zimek},
  title        = {Randomized {PCA} Forest for Outlier Detection},
  journal      = {CoRR},
  volume       = {abs/2508.12776},
  year         = {2025}
}

@article{DBLP:journals/ai/KohaviJ97,
  author    = {Ron Kohavi and
               George H. John},
  title     = {Wrappers for feature subset selection},
  journal   = {Artificial Intelligence},
  volume    = {97},
  number    = {1-2},
  pages     = {273--324},
  year      = {1997},
  url       = {https://doi.org/10.1016/S0004-3702(97)00043-X},
  doi       = {10.1016/S0004-3702(97)00043-X}
}

@inproceedings{DBLP:conf/icmla/EscanillaHKKSP18,
  author    = {Nicholas Sean Escanilla and
               Lisa Hellerstein and
               Ross Kleiman and
               Zhaobin Kuang and
               James D. Shull and
               David Page},
  title     = {Recursive Feature Elimination by Sensitivity Testing},
  booktitle = {17th IEEE International Conference on Machine Learning and Applications,
               ICMLA 2018, Orlando, FL, USA, December 17-20, 2018},
  pages     = {40--47},
  publisher = {IEEE},
  year      = {2018},
  url       = {https://doi.org/10.1109/ICMLA.2018.00014},
  doi       = {10.1109/ICMLA.2018.00014}
}
\end{document}